\documentclass[10pt,twocolumn,letterpaper]{article}

\usepackage[pagenumbers]{cvpr} 

\usepackage{graphicx}
\usepackage{amsmath}
\usepackage{amssymb}
\usepackage{url}            
\usepackage{booktabs}       
\usepackage{amsfonts}       
\usepackage{nicefrac}       
\usepackage{microtype}      
\usepackage{xcolor, color, colortbl}         
\usepackage{comment}
\usepackage{graphicx}
\usepackage{wrapfig}    
\usepackage{siunitx}  
\usepackage{xspace}
\usepackage{tabularx}
\usepackage{threeparttable} 
\usepackage{rotating}
\usepackage{multirow}
\usepackage{subcaption}
\usepackage{makecell}
\usepackage[export]{adjustbox}
\usepackage{enumitem} 
\usepackage{booktabs}
\usepackage{graphicx}

\usepackage{xcolor, color, colortbl}         
\definecolor{Highlight}{HTML}{39b54a}  

\usepackage{amssymb}
\usepackage{pifont}
\usepackage{algorithm}
\usepackage{listings}
\usepackage{etoolbox}
\makeatletter
\AfterEndEnvironment{algorithm}{\let\@algcomment\relax}
\AtEndEnvironment{algorithm}{\kern2pt\hrule\relax\vskip3pt\@algcomment}
\let\@algcomment\relax
\newcommand\algcomment[1]{\def\@algcomment{\footnotesize#1}}
\renewcommand\fs@ruled{\def\@fs@cfont{\bfseries}\let\@fs@capt\floatc@ruled
  \def\@fs@pre{\hrule height.8pt depth0pt \kern2pt}%
  \def\@fs@post{}%
  \def\@fs@mid{\kern2pt\hrule\kern2pt}%
  \let\@fs@iftopcapt\iftrue}
\makeatother
\newcommand{\cmmnt}[1]{}

\usepackage[normalem]{ulem}

\usepackage{listings}
\usepackage{color}
\definecolor{codegreen}{rgb}{0,0.6,0}
\definecolor{codegray}{rgb}{0.5,0.5,0.5}
\definecolor{codepurple}{rgb}{0.58,0,0.82}
\definecolor{backcolour}{rgb}{1,1,1}
 
\lstdefinestyle{mystyle}{
    backgroundcolor=\color{backcolour},   
    commentstyle=\color{codegreen},
    keywordstyle=\color{magenta},
    numberstyle=\tiny\color{codegray},
    stringstyle=\color{codepurple},
    basicstyle=\footnotesize,
    breakatwhitespace=false,         
    breaklines=true,                 
    captionpos=b,                    
    keepspaces=true,                 
    numbers=left,                    
    numbersep=5pt,                  
    showspaces=false,                
    showstringspaces=false,
    showtabs=false,                  
    tabsize=2
}
 
\makeatletter

\newcommand{\Rmnum}[1]{\expandafter\@slowromancap\romannumeral #1@}
\makeatother  

\usepackage[pagebackref,breaklinks,colorlinks]{hyperref}

\usepackage[capitalize]{cleveref}
\crefname{section}{Sec.}{Secs.}
\Crefname{section}{Section}{Sections}
\Crefname{table}{Table}{Tables}
\crefname{table}{Tab.}{Tabs.}

\usepackage{lipsum}

\def\cvprPaperID{****} 
\def\confName{CVPR}
\def\confYear{2023}

\begin{document}


\title{Estimating more camera poses for ego-centric videos is essential for VQ3D}


\author{Jinjie Mai \and Chen Zhao \and Abdullah Hamdi \and Silvio Giancola \and Bernard Ghanem\\
King Abdullah University of Science and Technology (KAUST) \\
{\tt\small \{jinjie.mai,chen.zhao,abdullah.hamdi,silvio.giancola,bernard.ghanem\}@kaust.edu.sa}}
\maketitle

\begin{abstract}
    Visual queries 3D localization (VQ3D) is a task in the Ego4D Episodic Memory Benchmark. Given an egocentric video, the goal is to answer queries of the form ”Where did I last see object X?” where the query object X is specified as a static image, and the answer should be a 3D displacement vector pointing to object X. However, current techniques use naive ways to estimate the camera poses of video frames, resulting in a low query with pose (QwP) ratio, thus a poor overall success rate. We design a new pipeline for the challenging egocentric video camera pose estimation problem in our work. Moreover, we revisit the current VQ3D framework and optimize it in terms of performance and efficiency. As a result, we get the top-1 overall success rate of 25.8\% on VQ3D leaderboard, which is two times better than the 8.7\% reported by the baseline.
\end{abstract}

\section{Introduction}
\label{sec:intro}

Recently, the large-scale video dataset Ego4D~\cite{grauman2022ego4d} was released, which contains more than 3,000 hours of ego-centric videos. As one of the Ego4D challenges, given questions like ”where did I last see object X?”, Visual Query 3D Localization (VQ3D) is the task to localize the 3D location of the object in first-person view videos, where the object query is an image crop. As a comparison, a lot of works have been done in Embodied Question Answering (EQA)~\cite{das2018embodied,datta2022episodic}, as a special case of video language grounding task, where an embodied agent should answer language questions according to visual observations in 3D indoor environment.

While EQA usually requires the model to give various answers~\cite{grauman2022ego4d} (e.g., language, video clip, etc.) to language query, VQ3D assumes images as query and predicts numeric object coordinates as output, which is more intuitive and fundamental for present computer vision techniques. VQ3D naturally requires our methods to perform robust pose estimation for ego-centric videos and relates 2D images and 3D scenes tightly in episodic memory, which is critical for 3D understanding, like the emerging embodied AI tasks. However, few works are done yet before the release of Ego4D.
In this paper, we revise VQ3D task at first. Then we give a formal definition of VQ3D task and clarify its metrics. We then discuss and analyze the baseline method. We find the main problem of the baseline is they failed to get enough camera poses estimation for ego-centric videos. To address this issue, we customize COLMAP to do a sparse reconstruction for the video and then register it back to the world coordinate system. The improved pose estimation module gives us over 50\% improvement in QwP, which yields a boost of 15\% in overall success rate.

\section{Background}

\begin{figure}[h]
\centering
\includegraphics[width=0.5\textwidth]{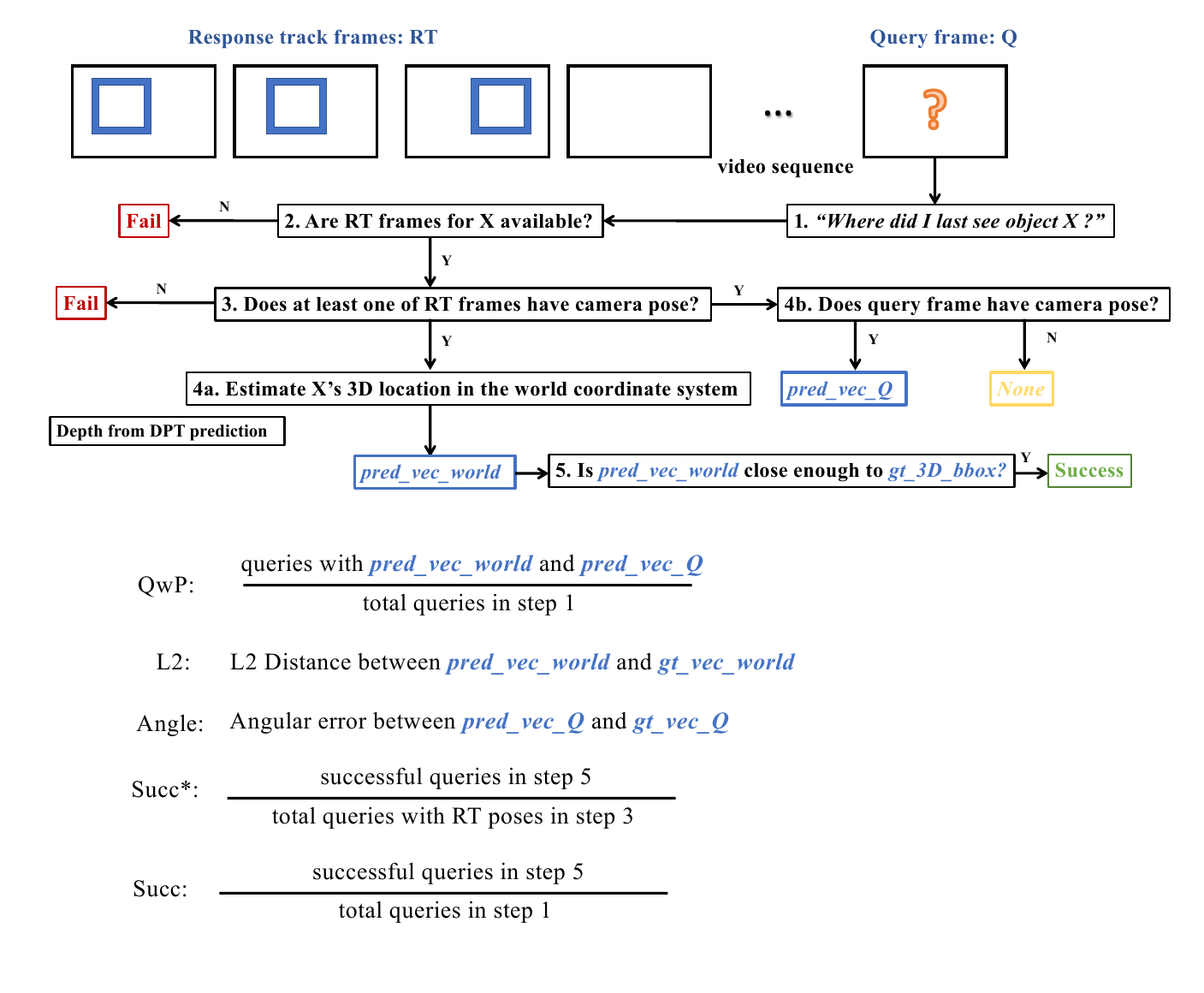}
\caption{Baseline pipeline and metrics}
\label{fig:main}
\end{figure}

Formally, given a visual query of the object $o$ specified by image $I$,
the task is to find the most recent 3D location of $o$ from 
query frame $q$.

The baseline method follows a multi-stage pipeline.
First, they estimate the camera intrinsics $K$ and poses for all the frames in video clip $v$.
Then, they use a pretrained VQ2D model~\cite{grauman2022ego4d,xu2022negative} to get a response track $r$,
which is a temporally contiguous set of frames that $o$ keeps appearing.
Next, they retrieve the most recent frame $f$ in $r$ with known camera poses $P_f$.
With the depth $d$ predicted by pretrained DPT network~\cite{Ranftl2021},
the 3D world location of $o$ is calculated by:
\begin{align}
    pred\_vec\_world=&
\begin{bmatrix}
x\\
y\\
z
\end{bmatrix}\\
    =P_fdK^{-1}&
\begin{bmatrix}
u\\
v\\
1
\end{bmatrix}
\end{align}
where $u,v$ is the center pixel coordinate in $f$ 
of the bounding box detection of $o$.

To elaborate on the pipeline and metrics adopted for this VQ3D challenge baseline,
we detail how 
predicted object location in world coordinate system $pred\_vec\_world$,
predicted object location in query frame system $pred\_vec\_Q$.
the ratio of queries with poses for both query frame and response track $QwP$,
L2 error between prediction and groundtruth annotation,
success rate $Succ^*$ computed only for queries with associated pose estimates,
and overall success rate $Succ$ with respect to all the queries are calculated 
in Fig \ref{fig:main}.



 




\section{Methodology}

\begin{table*}[htb]
\begin{tabular}{lcccccclc}
\toprule
                       & \textbf{Succ(\%)↑}   & \textbf{Succ*(\%)↑} & \textbf{L2↓} & \textbf{angle↓} & \textbf{QwP(\%)} & \textbf{RT} & \textbf{Pose Estimation} &  \\ \midrule
Baseline(Reproduction) & 8.71            & \textbf{51.47}           & \textbf{4.93}        & 1.23         & 15.15      & Last        & Baseline            &  \\
Ours                   & 22.73          & 40.12         & 8.27     & \textbf{1.19}         & 56.06      & Last        & SeqCOLMAP              &  \\
Ours                   & 21.59          & 38.79         & 13.84      & 1.24          & 56.44      & Average     & SeqCOLMAP           &  \\
Ours                   & 22.35          & 40.00              & 8.35       & 1.24         & 56.44      & Median      & SeqCOLMAP              &  \\
Ours                   & \textbf{25.76} & 38.74           & 8.97       & 1.20 & \textbf{66.29}      & Last        & SeqCOLMAP w/o filtering         & \\ 
\bottomrule
\end{tabular}
\caption{Results in test set. Note that all of these results are based on the prediction of baseline VQ2D model  }
\label{tb:main}
\end{table*}


\subsection{Poor $QwP$ found in baseline}
We carefully revisited the entire baseline pipeline as shown in Sec.~\ref{sec:intro}.
During our experiments, we found that the VQ3D performance is highly constrained by the ratio of queries 
we have poses $QwP$.
Actually, we can directly infer that $QwP$ is the upper bound of
the most important metric, the overall success rate 
from Fig.~\ref{fig:main}.
From the baseline results reported in Tb.~\ref{tb:main},
$QwP$ is as low as $15.12\%$ in the Ego4D baseline.
Therefore, the overall success rate is only $8.75\%$.

The main reason is that the baseline doesn't get enough camera poses for all the video frames.
In their implementation, the camera pose estimation is a four-step solution.
\begin{enumerate}
    \item \textbf{Camera Intrinsic Estimation.} They subsample a set of contiguous non-blurry frames from the video selected by the variance of Laplacian, which will be fed into COLMAP~\cite{schoenberger2016sfm} auto reconstruction to estimate camera intrinsics.
    \item \textbf{Keypoint extraction and matching.} They use SuperGlue~\cite{sarlin20superglue} to extract keypoints from all the video frames and match keypoints precomputed from Matterport 3D~\cite{Matterport3D}.
    \item \textbf{Coarse Came Poses.}  They use PnP from OpenCV~\cite{itseez2015opencv} library to solve the camera poses for the video frames that have at least 20 matched keypoints. Note that the 3D positions of 2D keypoints are computed by known intrinsics, depth, and poses from Matterport SDK.
    \item \textbf{Fine Camera Poses.} After matching between video frames and Matterport Scan, by way of matching between localized video frames and non-localized, they try to increase the number of valid estimates. In the end, they filter out those inaccurate pose estimations whose reprojection errors are larger than a threshold.
\end{enumerate}
However, we find the following issues in their implementation, resulting in a very low $QwP$: (1) In many videos, most frames don't have enough matches in Step 2. One reason could be the Matterport Scan, and the video is recorded at a different time. So they are under different environmental lighting, and the volunteers may change the scene's appearance in their daily activities. (2) In many video clips, the last step of filtering actually reduces the number of effective pose estimates instead of increasing it. Therefore, we think current filtering threshold is not the optimal setting.

\subsection{Improved pose estimation}
\subsubsection{Customized COLMAP for ego-centric videos}
To solve the issues we mentioned above, we propose to use SfM techniques to estimate the poses among the video frames directly to prevent poor matching performance in Step 2. Not only do we no longer need to worry about the inconsistency between scanned frames and video frames, but we also get rid of the strong assumption that there has to be an existing 3D scan of the video scene.

Specifically, we adopt the widely popular COLMAP~\cite{schoenberger2016sfm} as our tool for SfM and pose estimation. However, we find that the execution time of COLMAP increases exponentially if we use the exact auto reconstruction in Step 1 of the baseline. For a video clip consisting of 2,000 frames, the runtime of COLMAP can be up to a week. Because the default exhaustive matcher and dense reconstruction are very time-consuming though they provide good reconstruction quality. Such a speed is nearly unacceptable to finish pose estimation for the total 113 clips in VQ3D validation and test set.

To this end, we configure COLMAP to fit it in the egocentric video pose estimation scenario.
\begin{enumerate}
    \item We use separate commands to perform feature extraction, feature matching, and sparse reconstruction. Since we only need COLMAP to help us with pose estimation, we think it's reasonable to do sparse reconstruction only.  
    \item Because video frames are consecutive, we use \textit{SIFT feature extractor} and \textit{sequential matcher} for feature extraction and matching. This can significantly speed up the feature matching step.
    \item We configure the following parameters in Lst.\ref{lst:colmap} when we run COLMAP mapper. The motivation is to speed up the SfM further through a larger step for frame registration and a less number of global bundle adjustment.
    
 \begin{lstlisting}[language=Python, caption=Mapper configuration, label={lst:colmap}]
    'colmap', "mapper", 
    "--Mapper.ba_global_max_num_iterations", "30",
    "--Mapper.ba_global_images_ratio", "1.4",
    "--Mapper.ba_global_max_refinement", "3",
    "--Mapper.ba_global_points_freq", "200000"
\end{lstlisting}

We empirically find that, on average, we can obtain more than $90\%$ of the video frame pose estimates in one day with a 32-core CPU for a single clip. This is because it enables us to have enough good camera poses.

\end{enumerate}
 
\begin{figure}[h]
\centering
\includegraphics[width=0.5\textwidth]{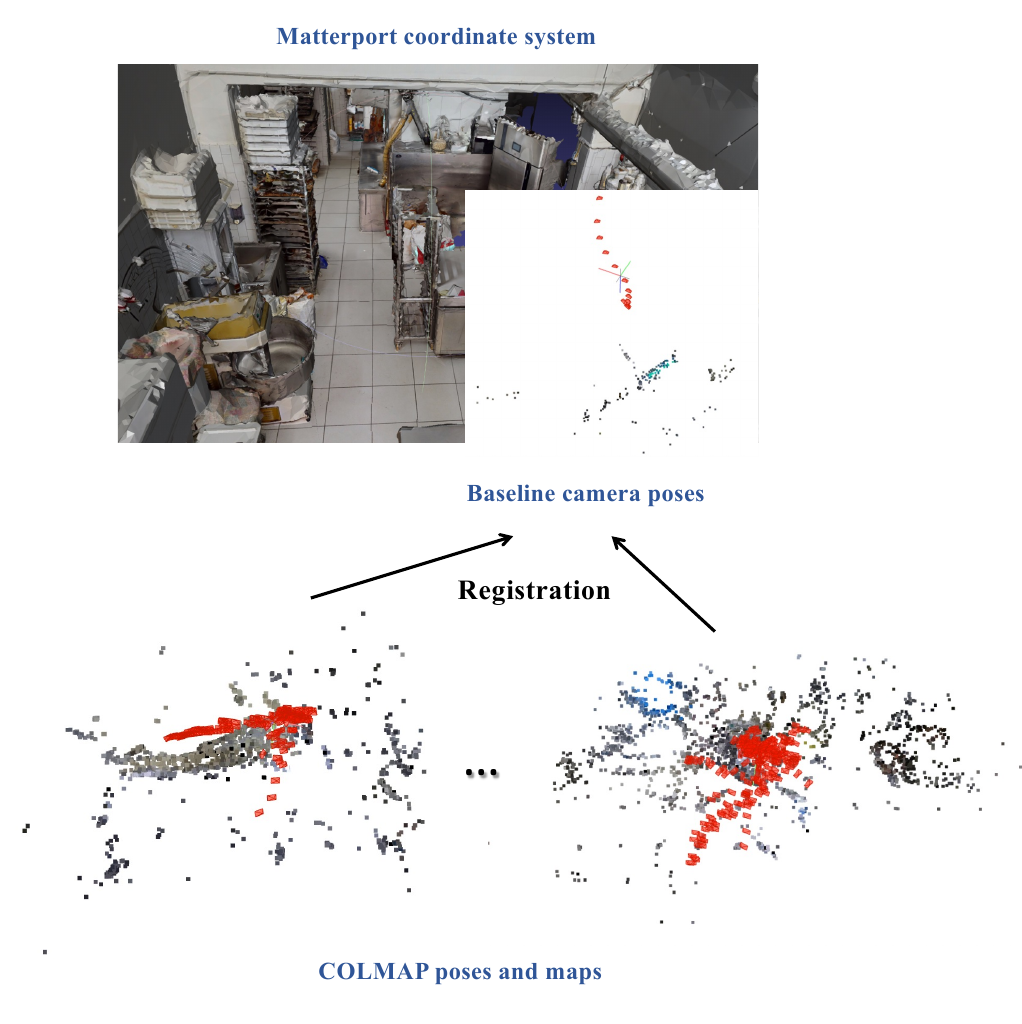}
\caption{COLMAP reconstruction and registration}
\label{fig:colmap}
\end{figure}

\subsubsection{Registration to the world coordinate system}
The problem with the poses from COLMAP is that they are independent of Matterport world coordinate system. We need to align our poses with the Matterport scan because volunteers make the groudturth annotations in those scans. 
We use the poses from the baseline method to do the alignment for time-being.

Assuming that we have two poses set $S_C$ and $S_M$ from coordinate system COLMAP and Matterport, respectively, we find the common frames set $S_CM$. Since we have poses in two systems for the same frames, we select the pose with minimal transformation error as the optimal transformation matrix
$T^*=[R|t]$ between $S_C$ and $S_M$. Also, as egocentric videos are always blurry and with abrupt motion,
COLMAP sometimes will lose track and construct multiple submaps with multiple pose sets $S_{C_1}, S_{C_2}....S_{C_n}$. As shown in Fig.\ref{fig:colmap}, we simply register $S_{C_{1,2...,n}}$ back into Matterport coordinate system one by one.

In $24$ of the total $69$ video clips, however, the baseline method can't find any good poses in all frames. Then we find that in some of these fail cases,
the baseline has some valid estimations in Step 3, but all of them are rejected in the filtering of Step 4. We then remove this threshold, which enables us to do the re-alignment again in $15$  of these $24$ fail cases. Removing the threshold is for sure not the optimal setting, but we always decide to try to have more camera poses even if they might not be accurate enough.

\section{Results}
\subsection{Experiment results}
Our experiment results and ablation study are shown in Tb.~\ref{tb:main}.
Note that we only finetune the camera pose estimation while keeping other steps of the baseline pipeline unchanged.
And the results are very impressive.
With customized COLMAP,
our overall success rate on Ego4D VQ3D benchmark is $22.73\%$, which improves more than $14\%$ on the $8.7\%$ reported by the baseline.
This vast improvement is brought by our much higher $QwP$, from just $15.15\%$ to $56.44\%$.
Furthermore, when we remove the filtering threshold to get more poses,
our method achieves the best result in the overall success of $25.76$,
another boost of $3\%$.
We can observe a trade-off between more poses and more minor L2 errors in our method.
But given that the bottleneck of VQ3D is insufficient poses,
we decide to have more pose estimations while scarifying some L2 performance.

\subsection{Ablation Study}
Moreover, we ablate on how to get predicted world coordinate $pred\_vec\_world$. We have tried to get $pred\_vec\_world$ by taking the average or the median of multiple frames in the response track. But we find the best result is obtained when we select the last frame of the response track.

\begin{figure}[t]
    \centering
    \includegraphics[width=0.4\textwidth]{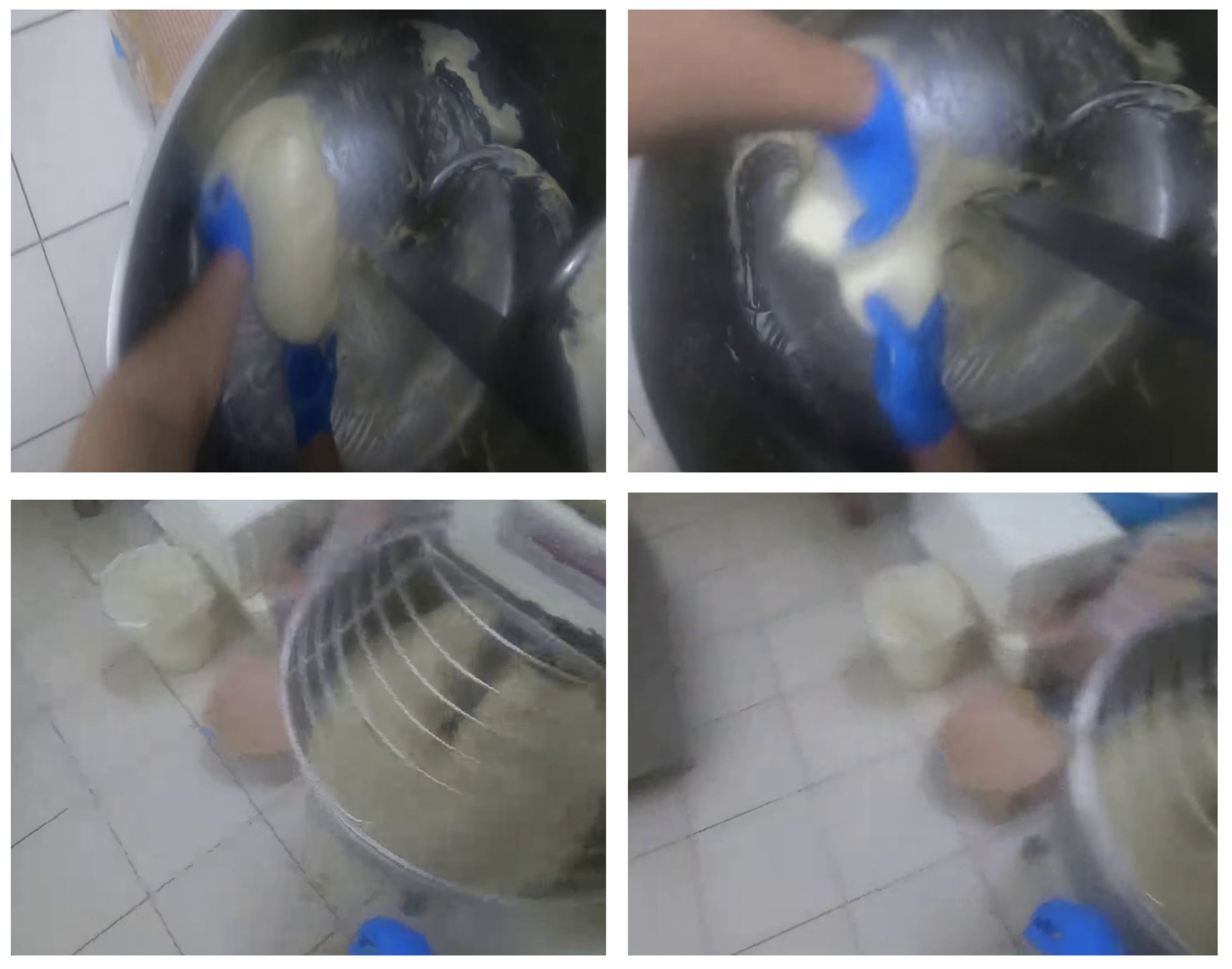}
    \caption{
        Example frames in a video clip we fail to get any pose estimation
    }
    \label{fig:fail}
\end{figure}

\subsection{Study on Failure Cases}
Since VQ3D challenge doesn't release the Matterport panoramas because of data anonymization concerns,
we have difficulties visualizing the objects in 3D.
Here we give some failed cases of camera pose estimation in Fig.~\ref{fig:fail}.
In this clip, the volunteer moves too fast, too close to the scene.
And the scene is dynamic and highly blurry because of the abrupt motion of the volunteer.

In conclusion, it's still quite challenging to estimate camera pose in the egocentric video.
The best $QwP$ from our finetuned COLMAP is $66.29\%$,
and there's still plenty of room for improvement.

\section{Limitations}
Though we make remarkable improvements on VQ3D benchmark, we think that there are still so many potential limitations. For example, we are still dependent on VQ2D results. Our current solution for camera poses is not accurate enough and can certainly be optimized further. Although our team achieves top-1 performance in the VQ3D public leaderboard now, we’ll study these problems and continue to push forward since VQ3D task is still far from resolved.

\section{Acknowledgement}
We thank Guohao Li, Jesus Zarzar and Sara Rojas Martinez for the discussion.
We also thank the Ego4D authors, Mengmeng Xu and Vincent Cartillier,
for their detailed explanation when we were trying to reproduce the VQ challenge baselines.

{\small
\bibliographystyle{ieee_fullname}
\bibliography{egbib}
}

\end{document}